\documentclass[conference]{IEEEtran}
\IEEEoverridecommandlockouts
\usepackage{cite}
\usepackage{amsmath,amssymb,amsfonts}
\usepackage{algorithmic}
\usepackage{graphicx}
\usepackage{textcomp}
\usepackage{xcolor}
\usepackage{soul}
\usepackage{url}
\usepackage{multirow}
\def\BibTeX{{\rm B\kern-.05em{\sc i\kern-.025em b}\kern-.08em
    T\kern-.1667em\lower.7ex\hbox{E}\kern-.125emX}}

\usepackage{hyperref}

\begin{document}

\newcommand{\linebreakand}{%
  \end{@IEEEauthorhalign}
  \hfill\mbox{}\par
  \mbox{}\hfill\begin{@IEEEauthorhalign}
}

\title{Snakes AI Competition 2020 and 2021 Report
}

\author{
\IEEEauthorblockN{Joseph Alexander Brown
\href{https://orcid.org/0000-0002-6513-4929}{\includegraphics[height=1.7ex]{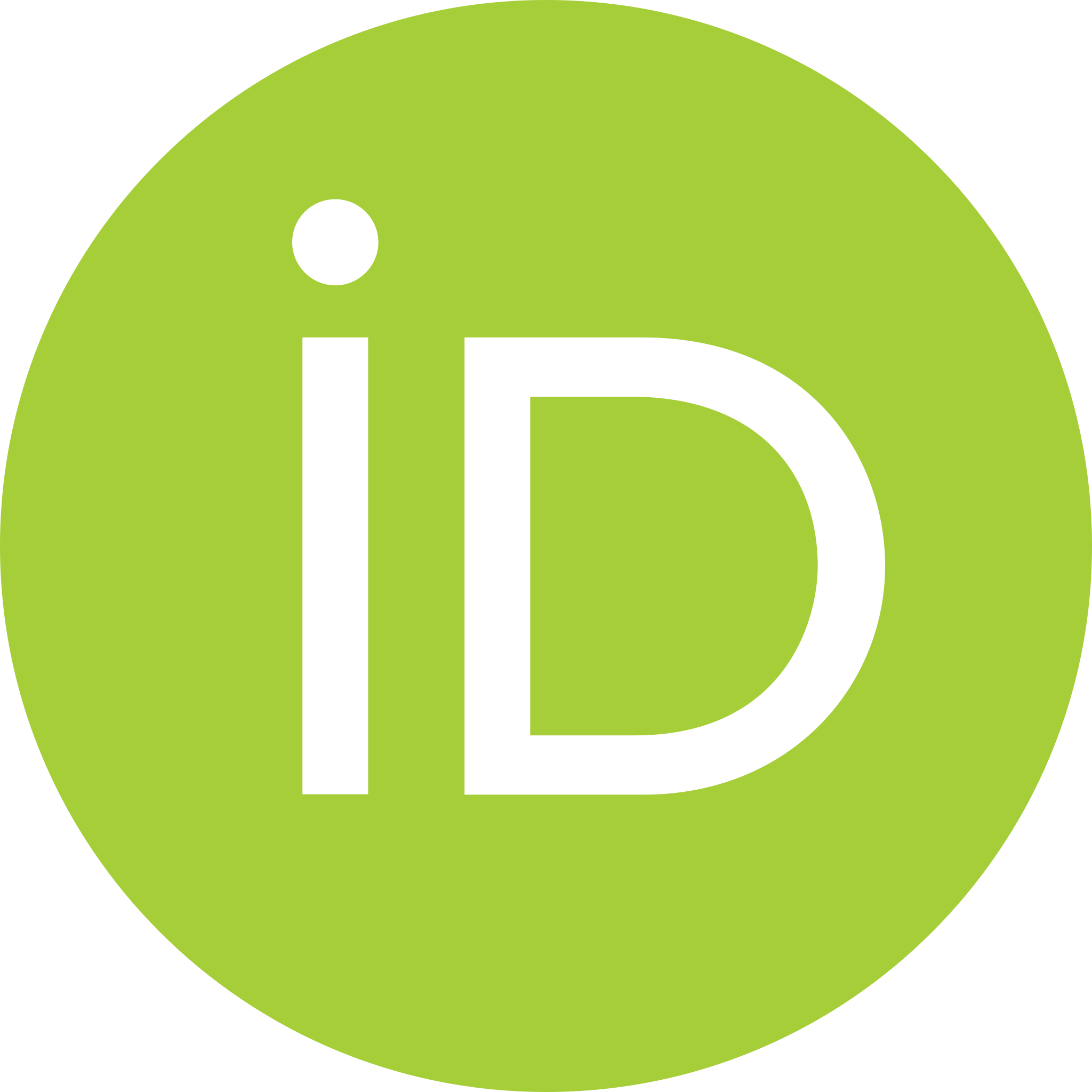}}}

\IEEEauthorblockA{\textit{AI in Games Development Lab} \\
\textit{Innopolis University}\\
Innopolis, Russia \\
j.brown@innopolis.ru}
\and
\IEEEauthorblockN{Luiz Jonat{\~a} Pires de Ara\'ujo \href{https://orcid.org/0000-0001-7450-7945}{\includegraphics[height=1.7ex]{images/ORCID_iD.svg.png}}} 
\IEEEauthorblockA{\textit{Innopolis University}\\
Innopolis, Russia \\
l.araujo@innopolis.university~\\~\\}
\and
\IEEEauthorblockN{Alexandr Grichshenko}
\IEEEauthorblockA{\textit{Innopolis University}\\
Innopolis, Russia \\
a.grishchenko@innopolis.university}
}

\maketitle

\begin{abstract}
The Snakes AI Competition was held by the Innopolis University and was part of the IEEE Conference on Games 2020 and 2021 editions. It aimed to create a sandbox for learning and implementing artificial intelligence algorithms in agents in a ludic manner. Competitors of several countries participated in both editions of the competition, which was streamed to create a synergy between organizers and the community. The high-quality submissions and the enthusiasm around the developed framework create an exciting scenario for future extensions.
\end{abstract}


\section{Introduction}

Our competition focuses on reestablishing Snakes, one of the most iconic cross-platform games, in a new light. The participants had the opportunity to delve into the familiar world of eating apples and avoiding their tails. Still, this time they were forced to quarrel against each other in an intense battle of wits.

The prototypical Snake game was \emph{Blockade} developed in 1976 as a two-player arcade game in a dedicated upright cabinet \cite{Blockade}.  In this classic arcade game, players' cursors started from predetermined and symmetrical positions and moved within a square making 90-degree turns. One of the difficulties players had to address was avoiding the running trail left behind their moving cursors. Hitting such trails would incur a loss for the infracting player. The movie Tron would also use these mechanics in the bike scenes, in which the bikes would leave a wall behind, causing a barrier to crash into for other participants \cite{Tron}.

Snake would become a single-player version of this add the idea of the walls left behind being limited in length and then encouraging the player to increase their score, and the risk of losing, via eating a power-up (known as an apple) to increase in length by a unit. This would serve as the foundation of all Snake games to come. Snake would be especially popularized via its inclusion on Nokia phones as a build-in monochromatic game \cite{Snake}. Our implementation for the contest game is closer to the \emph{Blockade} in that it is a two-player version but uses the apples from the \emph{Snake} variants to come.

This work is motivated by a more ludic environment to teach programming concepts and introduce artificial intelligence methods in university-level courses. Snakes were developed in 2019 during the course Introduction to Programming I at Innopolis University using provided base-code \cite{Araujo2019towards}. The course covers what the ACM /IEEE curriculum guidelines would specify as Tier-1 PL/SE \cite{ACM} and maps well to other European and North American first-year introductory classes in computer science. Students had the opportunity to develop the game and participate in a contest that rewarded students who implemented ingenious bots in a competition during the course. At the end of the semester, the instructors used the first Snakes match to reward students who developed ingenuous heuristics. The source code and example of bots are publicly available in GitHub\footnote{\url{https://github.com/BeLuckyDaf/snakes-game-tutorial}} and can be helpful resources for instructors willing to experiment with such an approach.

The implementation of this contest is based on Innopolis University's broad educational movement towards the implementation of active learning methods in the classrooms \cite{ASEE}. Active learning is known to have beneficial outcomes \cite{Freeman2014ActiveLI} but found to be hard to implement in computer science classes \cite{Eickholt2018BarriersTA}. Contests, competitions, and reflective exercises with students have been used as a method of increasing the active, practical learning of course projects \cite{Brown19Seeds,Brown19,Brown20}.

Retro-games have had a resurgence in the domain of AI agents, e.g. Pacman \cite{8207594} or Atari \cite{6756960}.  Their rule sets being simple enough to understand, their implementations being quick to program, and their time to run short, allowing for different evolutionary computation, neural training, or other learning systems to be used in this problem domain. Due to the simplicity, even those without this factor require little investment to understand the essential mechanics, or what a ``good'' player of the game's goal should be in terms of methods of winning or basic strategy. These games also benefit from their nostalgic factors for many of those researching them and those they are presented to, making them more accessible as research points.


The remainder of the paper is organized as follows: Section \ref{sec:rules} examines the rules of the Snake game utilized and Section \ref{sec:contestsystem} the implementation of the contest framework.  Sections \ref{sec:2020edition} and \ref{sec:2021edition} present the competitive teams and the contest results held as part of IEEE Conference on Games in 2020 2021, respectively, sponsored by the IEEE and Innopolis University. Section \ref{sec:conclusions} gives conclusions and future directions.

\section{Rules}
\label{sec:rules}

The contestants are required to implement in Java programming language their own AI agent, from now on referred to simply as bot. Each bot controls a snake to be the last standing snake or score the highest number of points in 3 minutes by eating apples. The rules are described as follows:

\begin{enumerate}
\item A bot controls only the direction the snake takes: north, south, east, or west.
\item Snakes always move simultaneously and forward. They also increase one position (i.e. pixel) after taking an apple.
\item A snake loses in any of these conditions occurs: if it leaves the board (blue snake in Figure \ref{fig:snakeleavestheboard}); if it hits its own body (white snake in Figure \ref{fig:snakehitsownbody}); if it hits the other snake's body (Figure \ref{fig:snakehitsopponnent}); if it takes more than one second to make a decision (i.e. which direction to take).
\item If the snakes collide head to head (Figure \ref{fig:headtohead}), then the longest snake wins. If the snakes have the same size, the game ends in a draw.
\item Apples appear randomly at an unoccupied position of the board, and there is only one apple available at any time.
\end{enumerate}

\begin{figure}[h]
  \centering
  \includegraphics[scale=0.35]{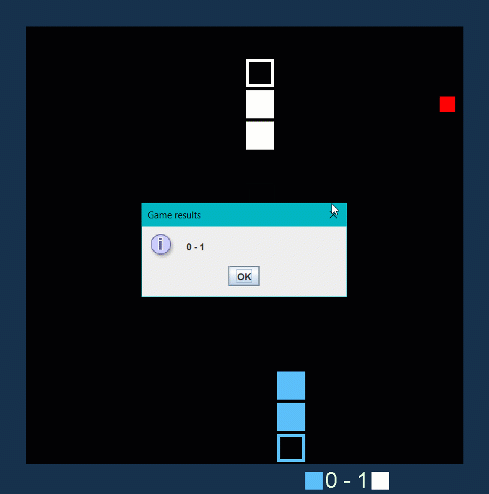}
  \caption{Blue snake leaves the board}
  \label{fig:snakeleavestheboard}
\end{figure}

\begin{figure}[h]
  \centering
  \includegraphics[scale=0.35]{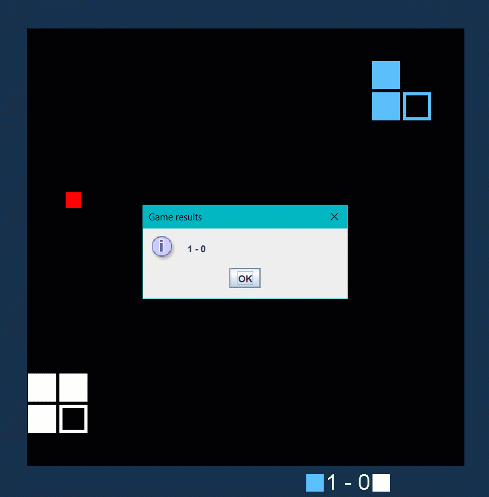}
  \caption{White snake hits its own body}
  \label{fig:snakehitsownbody}
\end{figure}

\begin{figure}[h]
  \centering
  \includegraphics[scale=0.35]{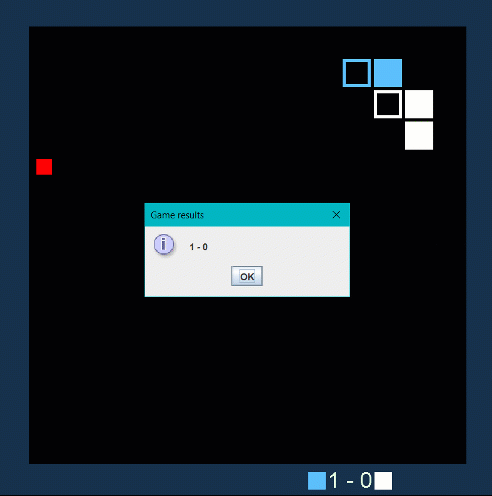}
  \caption{Snake hits the opponent's body}
  \label{fig:snakehitsopponnent}
\end{figure}

\begin{figure}[h]
  \centering
  \includegraphics[scale=0.35]{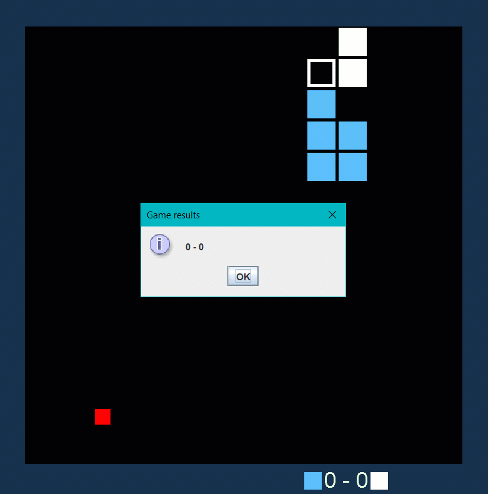}
  \caption{Head to head collision}
  \label{fig:headtohead}
\end{figure}

The following rule was added in the 2021 edition of the tournament: the apple disappears and then reappears at a random unoccupied position of the board if not taken after 10 seconds. The motivation for the introduction of this rule is described in Section \ref{sec:2020results}.

\section{Contest System}
\label{sec:contestsystem}

The format adopted for the AI Snakes Game Competition was the round-robin tournament (or all-play-all tournament). In other words, each contestant met all other contestants in turn. Moreover, each pair of challenging contestants played among themselves three times (30 in the 2021 edition) to mitigate randomness and reward more robust implementations. While a higher number of matches between the players would allow more statistically relevant data, it would also prevent a practical broadcast of the tournament online, mitigating the interest and consequently the iteration among the contestants.

In a match, the result for a particular contestant was either victory (1-0), draw (0-0) or defeat (0-1). AI bots were ranked at the highest number of wins, with ties broken with the highest number of draws.

\section{AI Snakes Game 2020 Edition}
\label{sec:2020edition}

\subsection{Competitors}

The following competitors participated in the 2020 edition:

\begin{itemize}
\item Eita Aoki, Heroz, Japan
\item Team \emph{Serpentine AI} \cite{serpentinereport}, Eindhoven University of Technology, Netherlands
    \begin{itemize}
    \item Bram J. Grooten
    \item Imre Schilstra
    \item Wolf van der Hert
    \item Dik van Genuchten
    \end{itemize}
\item Timo Bertram, University of Birmingham, United Kingdom
\item Vladislav Smirnov, Innopolis University, Russia
\item Danil Kabirov, Innopolis University, Russia
\item Alexey Zhuchkov, Innopolis University, Russia
\item Vyacheslav Vasiliev, Innopolis University, Russia
\end{itemize}

\subsection{Results}
\label{sec:2020results}

The competition was broadcast live, and it is available at 
\url{https://www.youtube.com/watch?v=eqxpSO2_HRI}. During the competition, the participants assessed their bots' performance, discussed the used algorithms, and observed some unexpected and inventive strategies. For example, when leading in the apple count, Serpetine's AI bot continuously shielded the apple to prevent the other bot from taking it, leading to victory after the timeout.

The log files with the results have been made publicly available at \url{https://agrishchenko.wixsite.com/snakesai}.
The final ranking sorted according to the number of victories is presented in Table \ref{tab:total}. Detailed results for each iteration can be found at the official website of the tournament.

\begin{table}[ht]
\centering
\caption{Final ranking}
\label{tab:total}
\begin{tabular}{|c|c|c|c|c|c|}
\hline
Participant & Wins \\
\hline
Eita Aoki & 24 \\ \hline
Danil Kabirov & 17 \\ \hline
Serprentine AI & 16 \\ \hline
Timo Bertram & 13 \\ \hline
Vyacheslav Vasiliev & 12 \\ \hline
Vladislav Smirnov & 6 \\ \hline
Alexey Zhuchkov & 5 \\ \hline
\end{tabular}
\end{table}

Common strategies and algorithmic techniques were shared among the top three contestants: iterative deepening search, apple encirclement, and survival prioritization. Some participants employed iterative deepening search (IDS) to navigate their snakes on the playing field. This approach combines the strengths of typical graph traversal algorithms breadth-first search (BFS) and depth-first search (DFS). This method was shown to optimize space utilization while delivering acceptable performance \cite{reinefeld1994enhanced}, which is also supported by the results of the competition. The top two players of the tournament employed variations of this approach.

Regarding the pure gameplay strategies, top players seemed to prioritize the survival of their snakes (i.e. avoiding collisions with opponents and staying within map bounds) over collecting apples. The plan is theoretically viable since as the snakes grow in size, the probability of collision increases, while the likelihood of obtaining the apple remains the same as there is only one apple on the map at any given time. An inventive strategy was employed by the \emph{Serpentine AI} team, whose bot effectively starved out the opponent by finding the apple and continuously move in circles around it, denying the opponent the opportunity to score a point. While such an approach was helpful in most cases, leading to higher winning scores, it is constrained to certain conditions. For example, the player should be fortunate enough to obtain the first couple of apples to justify the stalling.

In addition to the number of victories, running time is an exciting feature to analyze the contestants. Some bots utilize the full provided time (one second) to process the game state and decide. This caused some games to last longer and the rendering to be performed roughly. The trade-off between tolerating a long or short `thinking time' is the allowance or prevention of the use of more sophisticated AI algorithms and techniques. 


\section{AI Snakes Game 2021 Edition}
\label{sec:2021edition}

\subsection{Competitors}

The following competitors participated in the 2021 edition:

\begin{itemize}
\item Eita Aoki, Heroz, Japan
    \begin{itemize}
    \item Bot \#1: Thunder
    \item Bot \#2: Thunder 2020 (champion of the previous edition)
    \end{itemize}
\item Team \emph{Serpentine}, Eindhoven University of Technology, Netherlands
    \begin{itemize}
    \item Boris Muller
    \item Gijs Pennings
    \item Tunahan Sari
    \item Imre Schilstra
    \end{itemize}
\item Danil Kabirov, Innopolis University, Russia
\item Team Mike Python
    \begin{itemize}
    \item Andrew Boyley
    \item William Hill
    \end{itemize}
\item Team SOS, University of Tsukuba, Japan
    \begin{itemize}
    \item Kazuma Tokunaga
    \item Takuma Oishi
    \item Eiji Sakurai
    \item Daisuke Shimizu
    \end{itemize}
\item Evan Rex
\item Favour Obagbuwa
\item Iman Hosseini
\item Team Bishe Vora Nagin, Frankfurt University of Applied Sciences
    \begin{itemize}
    \item Munshi Fahim Sadi
    \item Md. R M Yusuf Naeem
    \item Rafayet Hossain
    \item Prithwika Banik
    \end{itemize}
\item Team Jordan, Frankfurt University of Applied Sciences
    \begin{itemize}
    \item Tobias Glotzbach
    \item Torben Heidt
    \item Ravinder Mahey
    \end{itemize}
\item Team Koffi, Frankfurt University of Applied Sciences
    \begin{itemize}
    \item Tobias Glotzbach
    \item Torben Heidt
    \item Ravinder Mahey
    \end{itemize}
\item Kevin Kopp, Frankfurt University of Applied Sciences
\item Abdelilah El Kalai, Frankfurt University of Applied Sciences
\end{itemize}

\subsection{Results}
\label{sec:2021results}

The competition run offline and lasted approximately 72 hours. The streaming of the results and discussion of strategies can be found at \url{https://agrishchenko.wixsite.com/aisnakes2021}.
The final ranking sorted according to the number of victories is presented in Table \ref{tab:total2021}. Detailed results for each iteration can be found at the official website of the tournament.

\begin{table}[h]
\centering
\caption{Final ranking}
\label{tab:total2021}
\begin{tabular}{|c|c|c|c|c|c|}
\hline
Team & Wins \\
\hline
Serpentine & 192  \\ \hline
Thunder2020 & 152 \\ \hline
Thunder & 147 \\ \hline
Team SOS & 138 \\ \hline
Mike Python & 124 \\ \hline
Team Jordan & 102 \\ \hline
Danil Kabirov & 98 \\ \hline
Evan Rex & 96 \\ \hline
Favour Obagbuwa & 82 \\ \hline
Team Koffi & 74 \\ \hline
XG - Iman Hosseini & 36 \\ \hline
Team Bishe Vora Nagin & 35 \\ \hline
Kevin Kopp & n/a  \\ \hline
Abdelilah El Kalai & n/a  \\ \hline
\end{tabular}
\end{table}

The last two places in the tournament standings are late submissions that did not participate in the entire running of the tournament. These submissions were added in a late mock competition with only three matches between every pair of players. The result is that the two late submissions ranked in the last two places.

Common strategies and algorithmic techniques were shared among the top three contestants: alpha-beta pruning and iterative deepening search, Monte Carlo Tree Search, and A* algorithm. As in the previous year, graph-based algorithms like BFS, IDS, alpha-beta pruning served as base strategies for the best-ranked participants. Moreover, evaluation functions were manually designed to assess the quality of problem states. For example, Serpentine's evaluation is a combination of the snake length, the distance to the apple, and the position of the snakes. The latter parameter considers the number of empty cells (after the apple is eaten) closer to their bot than the opponent. Other participants employed variations of this evaluation function.

After the announcement of the results, the contestants participated by giving suggestions for future editions of the tournament. The suggestions include:

\begin{itemize}
    \item Changing the game from two to three-dimensional elements and visualization. Figure \ref{fig:inspiration3dsnakes} presents an inspiration for intended interface.
    \item Adding obstacles to reduce the number of available moves at a time for a snake.
    \item Enabling the use of Python to implement the bots.
\end{itemize}

\begin{figure}[h]
  \centering
  \includegraphics[scale=0.4]{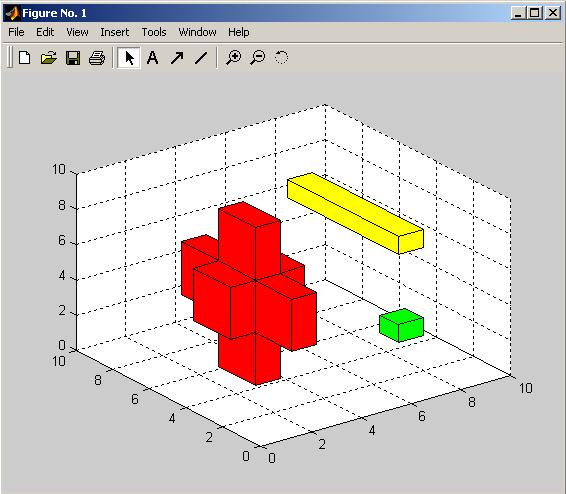}
  \caption{Inspiration for the visualization of the game in future editions. Source: https://www.mathworks.com/matlabcentral/fileexchange/3280-voxel.}
  \label{fig:inspiration3dsnakes}
\end{figure}


\section{Conclusions}
\label{sec:conclusions}

The Snakes AI Competition held by Innopolis University was a successful experience that allowed a sandbox environment to apply AI algorithms in a ludic manner. This is part of a movement towards using more active learning methods known to provide better student outcomes. Participants of several countries participated and could enjoy the competition online, interacting with organizers and other contestants in a friendly and educational environment.

Participants implemented techniques that employ mainly graph algorithms and human-designed heuristics. In the future, practices such as Genetic Programming (GP) and Artificial Neural Networks (ANN) will be encouraged in their use in class or future competitions. Future research will address the use of GP for evolving evaluation functions that can aid the search and pruning of the search tree of game states.

\section*{Acknowledgements}
The authors would like to thank IEEE for providing a \$500 prize for the winner of the 2020 edition of the tournament.

The authors also thank the community members who contributed with corrections and improvement of the source code on GitHub.

Finally, we thank instructors and students from the Frankfurt University of Applied Sciences for their active participation in the competition.

\bibliographystyle{IEEEtran}
\bibliography{snakes.bib}

\end{document}